\documentclass[11pt,a4paper]{article}
\usepackage[hyperref]{acl2021}
\usepackage{times}
\usepackage{latexsym}
\usepackage{graphicx}

\usepackage{microtype}

\aclfinalcopy % Uncomment this line for the final submission
\title{Doc2Dict: Information Extraction as Text Generation}

\author{
   Benjamin Townsend\footnotemark[1]~~ Eamon Ito-Fisher\footnotemark[2]~~ Lily Zhang\footnotemark[3]~~ Madison May\footnotemark[1] \\
   \footnotemark[1] Indico Data Solutions \\
   \footnotemark[2] Franklin W. Olin College of Engineering \\
   \footnotemark[3] New York University \\
   \texttt{\{ben,madison\}@indico.io} \\
   \texttt{efisher@olin.edu} \\
   \texttt{lily.h.zhang@nyu.edu}
}

\date{}

\begin{document}
\maketitle
\begin{abstract}

Typically, information extraction (IE) requires a pipeline approach: first, a sequence labeling model is trained on manually annotated documents to extract relevant spans; then, when a new document arrives, a model predicts spans which are then post-processed and standardized to convert the information into a database entry. We replace this labor-intensive workflow with a transformer language model trained on existing database records to directly generate structured JSON. Our solution removes the workload associated with producing token-level annotations and takes advantage of a data source which is generally quite plentiful (e.g. database records). As long documents are common in information extraction tasks, we use gradient checkpointing and chunked encoding to apply our method to sequences of up to 32,000 tokens on a single GPU. Our Doc2Dict approach is competitive with more complex, hand-engineered pipelines and offers a simple but effective baseline for document-level information extraction. We release our Doc2Dict model and code to reproduce our experiments and facilitate future work. \footnote{The code is released at \url{https://github.com/IndicoDataSolutions/Doc2Dict}.}

\end{abstract}

\begin{figure*}
    \centering
    \includegraphics[width=\textwidth]{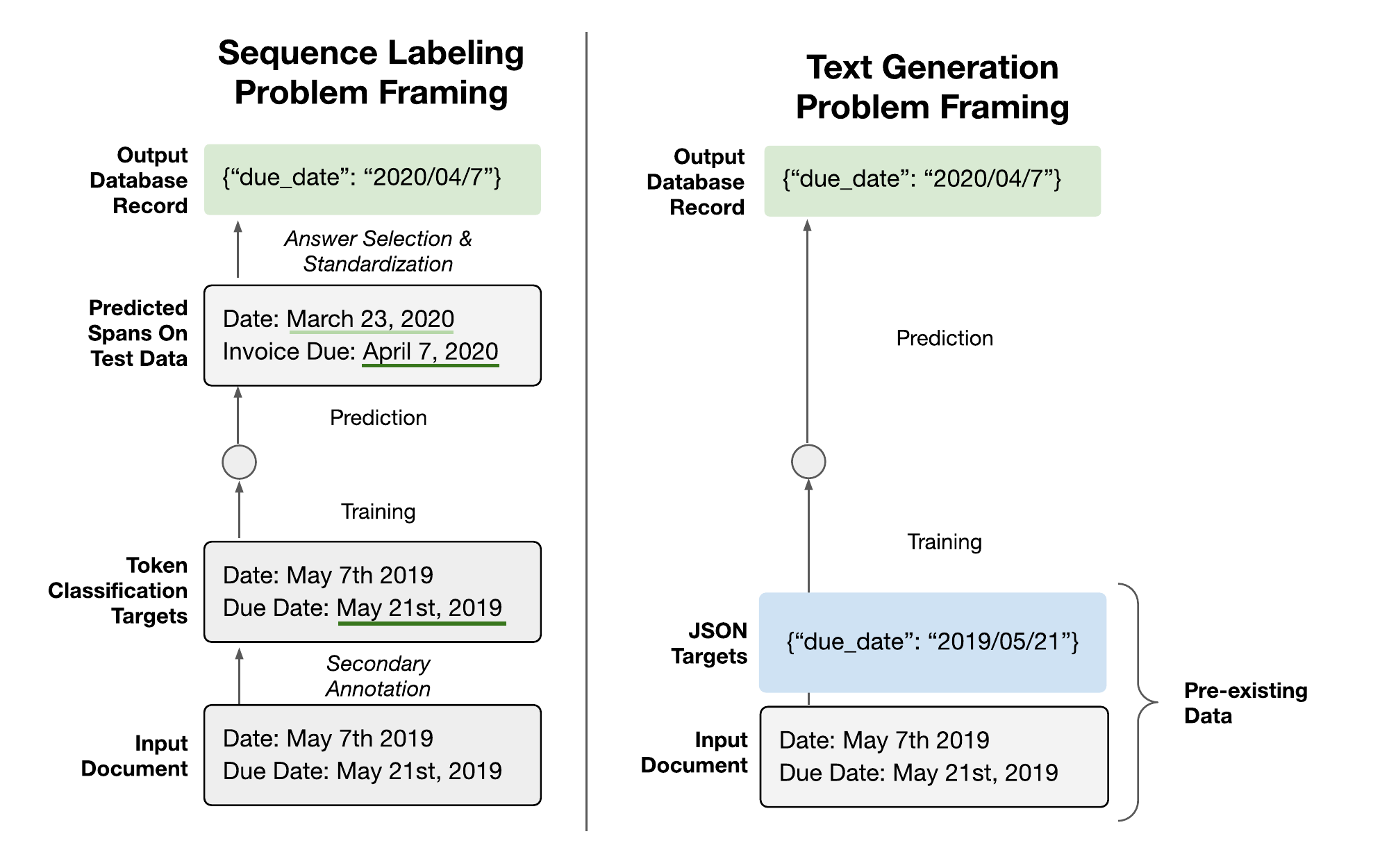}
    \caption{Information extraction with sequence-labeling models requires gathering additional token-level labels as well as an additional span selection and standardization step.  Our proposed method, Doc2Dict, learns to directly produce a normalized version of the database record that corresponds to a given document, enabling the use of pre-existing database records as supervision and minimizing the need for custom post-processing.}
    \label{fig:processing-steps}
\end{figure*}
\section{Introduction}
\label{introduction}
 The task of extracting structured information from documents is omnipresent \cite{Zhang2020, Bodhisattwa2020, Liu2019}. Although documents are designed primarily for human-consumption, they often contain values (e.g. amounts, dates, IDs) that are useful as inputs to downstream software. While rigid forms permit straightforward automated processing via the use of document templates, more varied documents like invoices, contracts, and quarterly reports pose a challenge for machine-learning assisted processing. For these documents, the information necessary to convert the document to structured records is primarily textual and we turn to the field of natural language processing for potential solutions.

\subsection{Sequence-Labeling for Information Extraction}
It is natural to frame information extraction problems as sequence-labeling tasks \cite{Collobert2011, Huang2015BidirectionalLM, lample-etal-2016-neural}. However, sequence-labeling models rely on the availability of token-level annotations — data not typically recorded by data entry software, which only captures the standardized values derived from the document. Consequently, documents that have already been manually processed need to undergo a secondary annotation process where annotators denote document spans that correspond to the derived values.

The cost of annotating large corpora with per-token labels is significant \cite{Alonso2016, Settles2008}, and this burden is exaggerated for long documents where information of interest may be spread throughout the document. Previous work has focused on minimizing the number of labeled samples required to construct an effective model \cite{Howard2018UniversalLM, Radford2019, Devlin2019}, but annotating even a few hundred lengthy documents still requires substantial effort and domain knowledge.

One might reasonably look to reconstruct token-level labels by automatically finding the source of each field in the corresponding document, but this alignment process is error prone. Searching for exact string matches between database values and source document text is insufficient, as values are frequently standardized prior to storage (e.g. ``May 7th, 2019'' may be converted to ``2019/05/07''). 

In other cases, a stored value might spuriously match irrelevant tokens in the source document. For instance, the ``Subtotal'' and ``Total'' lines on an invoice may both match the value of ``Invoice Total'' stored in a database, but only the latter should be provided as an annotation. More sophisticated systems do exist for aligning the values stored in a database with likely matches in the source document, but they rely on custom logic for each field type to perform reliably \cite{gralinski2020} and must be tailored to the task of interest.

Even after constructing a dataset with token-level labels and training a sequence-labeling model, it is necessary to post-process predicted spans to prepare entities for storage in a database. In the case that a single value is expected for a given field, a pool of candidate spans must be narrowed down to a single response via a selection heuristic.

Values must also be standardized to match the expected format. If standardization is handled by a rules-based system, additional software engineering effort is necessary in order to benefit from the machine learning system. If standardization is instead performed manually, extra overhead will be incurred in perpetuity when incoming documents are processed.  In combination, these limitations pose significant barriers to adoption.

\subsection{Information Extraction as Text Generation}
One possible solution to address the difficulty of producing labeled training data for sequence-labeling tasks is to forgo the requirement of token-level labels altogether by framing information extraction as a sequence-to-sequence task from document to a structured database record directly \cite{Sutskever2014}. Sequence-to-sequence learning presents a flexible framework under which many diverse tasks can be unified - classification, regression, question answering, and sequence-labeling tasks can all be treated as simple text generation tasks \cite{McCann2018, Raffel2020, Radford2019}. Rather than iterating through a document and predicting which document spans correspond to the fields of interest, we task a machine learning model with generating a JSON-like representation of the extracted information when conditioned on the text of the source document, permitting the direct use of preexisting database records as supervision.

This paper introduces Doc2Dict: a general, end-to-end information extraction system that directly utilizes database records generated by manual data entry processes as training data for a text generation model.  Doc2Dict requires little to no custom configuration and eliminates the need for additional manual annotation or a heuristic alignment step.  We jointly learn to extract, select, and standardize values of interest, enabling direct generation of a JSON-like data format. Additionally, we leverage gradient checkpointing to allow processing documents of up to 32k tokens on a single consumer GPU. Finally, we systematically explore how different types of information loss in the translation to database records impact model performance and suggest avenues for minimizing impact.
 
\section{Related Work}

Other natural language processing application areas have already benefited from the use of text-generation models to tasks that require structured outputs.  In 2013, \citet{andreas-etal-2013-semantic} demonstrated that framing semantic parsing as a machine translation task was competitive with existing methods that were purpose-built for semantic-parsing.  More recently, text-generation methods have been successfully applied to the text-to-SQL task, which requires the translation of a natural language question and a database schema to a SQL query. Of particular note is SeqGenSQL \cite{Li2020}, a finetuned T5 model that treats the text-to-SQL problem as a pure text generation task and currently leads the weakly supervised category of the WikiSQL benchmark \cite{zhongSeq2SQL2017}.  

Similar to the text-to-SQL problem, task-oriented dialogue systems often contain components that translate free-form conversations to structured representations of dialogue state.  While task-oriented dialogue systems have traditionally been broken up into many disparate parts (dialogue state tracking, dialogue management, and response generation), SimpleTOD \cite{HosseiniAsl2020ASL} and its successors \cite{Peng2020SOLOISTFT, lin2020mintl} unify these parts under the umbrella of text generation and use a simple language modeling objective to learn to produce dialogue state representations. 

Several prior works explore information extraction tasks in an end-to-end manner. \citet{Paolini2021} proposes an augmented natural language format for use in solving a wide variety of structured prediction language tasks, including named entity recognition, relation extraction, and event extraction. Their output format and framing of IE as a generative language modeling task permits multi-task learning without task-specific heads, but requires span-level information, preventing direct application to our task of interest. \citet{Zeng2018ExtractingRF} and \citet{Nayak2020EffectiveMO} use text-generation models to perform end-to-end joint entity and relation extraction from sentences.  Similarly, \citet{Lin2020AJN} jointly tackles entity, relation, and event extraction tasks by using a transformer encoder and sophisticated graph decoder. In all cases, however, copy mechanisms or extractive task formulations prevent straightforward application to our problem setting.

Most relevantly, a handful of prior works have approached information extraction from business documents in an end-to-end manner. \citet{Palm2019} leverages pointer networks to extract relevant fields from invoices. However, they build custom neural modules to handle the standardization of each field type, which limits the general usability of the approach to other output types not yet considered (e.g. addresses) and prevents leveraging pre-trained models to minimize training data requirements. \citet{Sage2020} similarly employs a pointer-generator network for information extraction from purchase orders. In contrast to our work, like \cite{Zeng2018ExtractingRF} they assume that no standardization has been applied to the extracted values and require that all system outputs can be copied directly from the source document.

\section{Methods}
\label{methods}

Our Doc2Dict model is based on the T5 transformer-based encoder-decoder architecture of \cite{Raffel2020}. We maintain the text-to-text formulation of T5, enabling structured output generation by formatting the output as Python's string representation of built-in data types. In practice, we find that a multitude of different structured output formats enable similar performance (see Section \ref{output-format}). In the following subsections, we present additional considerations raised by our specific formulation of information extraction as text generation and our methods for addressing them. 

\subsection{Entity Order}
\label{sec:entity-order}

One peculiarity of treating information extraction tasks in a text-to-database formulation is that the correct output sequence is not uniquely determined; in particular, any ordering of key-value pairs is a correct output. Language models trained with teacher forcing, however, only consider one correct output string. Using such a loss, our model may be penalized for producing valid responses. Unfortunately, training on all possible permutations increases the computational requirements by a factor of $n!$ where $n$ is the length of the sequence. Instead, we choose to encourage robustness to entity order by evaluating with respect to different randomly-ordered output sequences per epoch. We find that this simple solution works reasonably well in practice, achieving gains over fixed output orderings.

\subsection{Standardization}
\label{section:record-standardization}
The standardization of output values results in a discrepancy between the supervision offered by database records and the text that exists in the document. For instance, if the entity of interest is a due date that exists in YY/MM/DD format in the database but in textual format (e.g. Month Day, Year) in the document, correctly predicting the provided database label requires that a model not only identify the relevant part of the text, but also translate that text into the format given in the label. The latter must be learned without explicit supervision of an intermediate value (e.g. the exact value as seen in the document) and often requires learning non-trivial multi-token to multi-token mappings. For instance, even translating between words with different capitalization can change tokenization drastically, e.g. (``Barack'', `` Obama'') vs (``B'', ``AR'', ``ACK'', `` '', ``OB'', ``AMA''). To alleviate this effect, we preprocess the input text before tokenization so that candidate text spans more closely align with the database output, e.g. if outputs are all lowercase, we would lowercase all the text before feeding it into the model. 

\subsection{Processing Long Input Sequences}
\label{section:issues-with-chunking}
Transformer-based language models are typically trained on subsequences of 512 to 1024 tokens at a time because of the memory complexity of the attention operation \cite{Vaswani2017AttentionIA, Wang2020LinformerSW}, but many information extraction tasks require processing documents that may be tens of thousands of tokens in length. In the sequence-labeling problem formulation, this limitation can be overcome by ``chunking'' the document into either sentences or windows that contain a fixed number of tokens \cite{Dai2019TransformerXLAL}. 

Unfortunately, this chunking paradigm breaks down when applied to a text-generation setting. Using each chunk and the target sequence independently is ill-formed, as many document chunks may contain no information relevant to producing the target sequence. If the information present in the target sequence is not derivable from the input subsequence supplied, the loss encourages the undesirable behavior of producing likely output sequences not relevant to the input. Without knowing where in the document target values are derived we cannot effectively break the document up into manageable subsequences with appropriate labels per chunk.

\subsubsection{Fusion-in-Decoder} 
As the aforementioned chunking scheme is not viable for our approach, we address the issue of long input sequences by taking inspiration from the fusion-in-decoder method \cite{Izacard2020LeveragingPR}.  First proposed for applications in open-domain question answering, fusion-in-decoder provides a means for encoder-decoder language models to independently encode many subsequences but aggregate information from all subsequences in the decoder. In \citet{Izacard2020LeveragingPR}, fusion-in-decoder is applied to different documents returned by a retriever, whereas we apply a similar method to permit conditioning on many chunks from a single long document. Rather than a quadratic dependency on overall sequence length, the fusion-in-decoder method requires $O(c^2m)$ compute and memory, where $c$ is the chunk size and $m$ is the number of fixed-size chunks processed by the encoder. By only permitting dense attention within a given chunk in the encoder and restricting chunks to a fixed size, the memory and compute complexity is made linear with the length of the document.  

\subsubsection{Gradient Checkpointing Scheme}
\begin{figure}[t!]
    \includegraphics[width=\columnwidth]{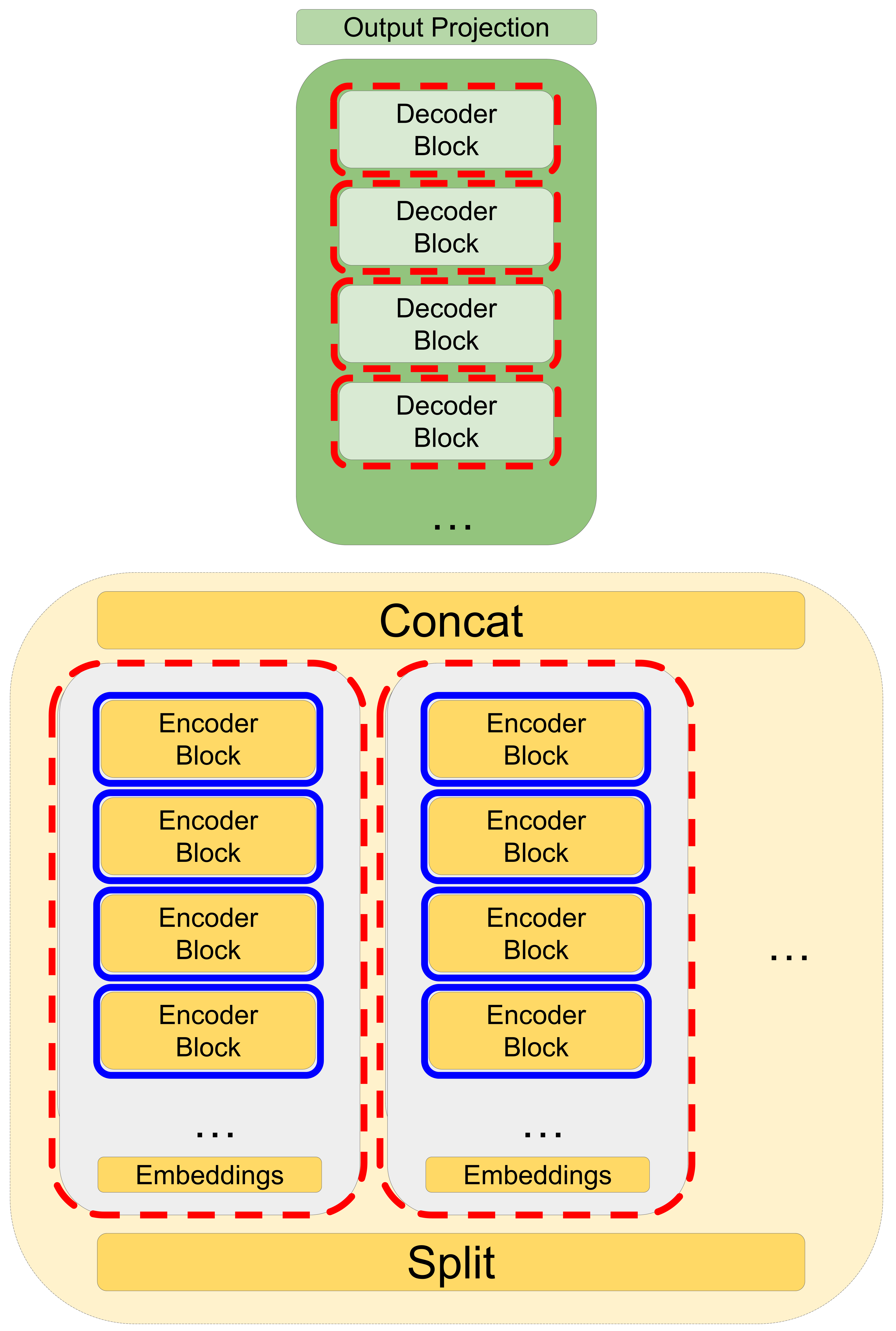}
    \caption{Gradient checkpointing scheme used to make training on long input sequences tractable. We compute a full forward pass of the model, computing the loss and caching the activations of each encoded chunk and each of the inter-block activations of the decoder (indicated by dashed red lines). These cached activations are used in conjunction with a second forward pass through the decoder in order to compute the gradients of the decoders parameters. For each block in the encoder, we then recompute the forward pass to cache the inter-block activations (indicated by solid blue lines).  A final forward pass through the encoder allows computing gradients with respect to the parameters of the encoder.}
    \label{fig:fusion-in-decoder}
\end{figure}

Even with these algorithmic optimizations, the memory requirements of processing long documents remain high and gradient checkpointing is required at training time. Gradient checkpointing allows trading compute for memory by caching only a portion of intermediate activations and recomputing the remaining activations during the backwards pass \cite{Chen2016TrainingDN}. We use this standard approach for the decoder, but for the encoder we add an additional layer of gradient checkpointing per-encoder stack, illustrated in figure \ref{fig:fusion-in-decoder}. When using a batch size of 1 at training time, this approach allows us to train on 64 blocks of 512 tokens at full precision with only 11GB of VRAM.  This translates to an overall sequence length of over 32,000 tokens.

\subsection{Output Format Experiments}
\label{output-format}
% Since we have elected to frame this information extraction task as a text-generation, we must decide on an output format to use in representing the content of the database records we use as supervision.  
To test what output formats are most easily generated by T5, we experiment with multiple output formats on a joint intent detection and slot-filling task, ATIS \cite{hemphill1990}. Output formats tested include XML, JSON, and Python's string representation, a close cousin to JSON. As the T5 vocab is missing some tokens necessary for representing these formats (e.g. ``\{'' and ``\}''), some of the extra IDs present in the T5 vocab were assigned to these tokens. 
 
% Plot performance by output format
Although the lack of the requisite tokens in the vocabulary implies that T5 has not been exposed to data in these formats during pre-training, we found that all output formats were viable targets for the task and performed similarly, as shown in Table \ref{tab:output-formats}. We opt to use Python's str format in subsequent experiments, due in part to its shorter output sequence length relative to XML. This shorter output length translates directly to shorter training and prediction times.

\begin{table}[t]

\centering
\begin{tabular}{llll}
\hline \textbf{Format} & \textbf{Intent} & \textbf{Entity} & \textbf{Parse Failure} \\ \hline
YAML & .970 & .951 & .003 \\
XML & .969 & .949 & .002 \\
JSON & .972 & .949 & .002 \\
Python string & .971 & .950 & .004 \\
\hline
\end{tabular}
\caption{\label{tab:output-formats} Performance on ATIS joint intent and slot filling task by output format. Output format used has negligible impact on task performance.  Reported intent and entity scores are micro F1. Across all output formats, generated content failed format validation only 0.26\% of the time.}
\end{table}

\section{Results}
We evaluate our method on a number of datasets covering direct extraction and end-to-end information extraction, the results of which are included in Tables \ref{tab:kleister-deepform-results}.

One of the primary strengths of our approach is its ability learn to extract and standardize data from documents given only the desired answer for training signal. We demonstrate these strengths on three datasets; Kleister-NDA, Kleister-Charity \cite{gralinski2020} and DeepForm \cite{DeepForm2020}. Each of these datasets provides a set of documents paired with ground truth values, but does not provide information about where ground truth values occurred in the source document. Some of these ground truth values appear verbatim whilst others have been transformed to a standardized format. For the Kleister and DeepForm datasets the baselines used in the original papers apply a pipeline approach, where custom heuristics are used to align records to text spans in the input document and re-cast the problem as a sequence-labeling task. At prediction time per-field standardization and selection rules are applied to the model predictions in order to translate extracted spans back to a canonical form that matches the desired output.

Additionally, we would like to draw some conclusions about how token-level supervision compares to a sequence-to-sequence formulation when entity standardization is not required. As an indicator of relative performance we compare a RoBERTa model trained with token-level annotations to our Doc2Dict method. We test on the ATIS intent detection dataset as well as the OntoNotes 5.0 NER dataset \cite{pradhan-etal-2013-towards}.

\subsection{Metrics}
For the Kleister datasets we use uncased F1 metrics to match the original task paper \cite{gralinski2020}. For DeepForm we use uncased entity-level F1 and report macro F1 as an overall score.  For ATIS and OntoNotes entity-level micro F1 is reported.

\subsection{Data Preprocessing}
For Kleister and DeepForm, we opt to lowercase both the source and the targets and remove any commas that occur directly after a number in the source text.  This is an effort to rectify some of the problems resulting from inconsistent tokenization, mentioned previously in Section \ref{section:record-standardization} and discussed at length in Section \ref{section:standardization}. The relevant metrics are insensitive to case so this transformation does not effect metrics. Ablations for these transformations can be seen in Table \ref{tab:kleister-ablations}.

For the DeepForm and ATIS datasets, our output format is a Python dictionary format.
\begin{verbatim}
{'key': value, ...}
\end{verbatim}
For the Kleister and OntoNotes datasets, our output format is a set of tuples, which allows us to represent keys that have multiple values.
\begin{verbatim}
{('key', value), ...}
\end{verbatim}

% TODO: figure out how to make this not run off the edge of the page or cut it
% \begin{verbatim}
% {'gross_amount': 59415.0, 'committee': 'OBAMA FOR AMERICA', 'agency': 'GREER, MARGOLIS, & MITCHELL #1', 'callsign': 'WOIO'}
% \end{verbatim}

\begin{table}[t]
    \centering
    \begin{tabular}{lrr}\hline
        Dataset & Doc2Dict F1 & Pipeline F1 \\
        \hline
        DeepForm &\textbf{0.900} &0.761 \\
        Kleister-NDA &\textbf{0.809} &0.777 \\
        Kleister-Charity &0.561 &\textbf{0.801} \\
        \hline
\end{tabular}
    \caption{Results for Kleister-NDA, Kleister-Charity and DeepForm.  In two of three tasks of interest, we outperform the baseline despite not requiring token-level supervision. Although we under-perform the baseline on Kleister-Charity, we believe that the comparative ease and reduced cost of obtaining the required training data has the opportunity to result in significant benefits for large scale practical applications.}
    \label{tab:kleister-deepform-results}
\end{table}
\begin{table*}[]
    \centering
    \begin{tabular}{lrr}
        \hline
        Doc2Dict &F1 &Baseline F1 \\
        \hline
        OntoNotes & 0.865 & \textbf{0.896} \\
        ATIS & 0.959 & \textbf{0.987} \\
        \hline

        % 64 epochs, batch size 8
    \end{tabular}
    \caption{The baseline model is a RoBERTa model \cite{Liu2019} with CRF output layer that has access to gold standard token-level annotations as training signal. Even without this additional supervision, our model is able to produce very reasonable results on OntoNotes and ATIS datasets.}
    \label{tab:ontonotes-atis-results}
\end{table*}

\begin{table}[]
    \centering
    \begin{tabular}{llrr}
        \hline
        Dataset & &F1 \\
        \hline
        Charity &Base &0.503 \\
        Charity &+ Lowercased &0.533 \\
        Charity &+ Comma Stripping &0.561 \\
        \hline
        NDA &Base &0.716 \\
        NDA &+ Shuffled Epochs &0.759 \\
        NDA &+ Lowercased &0.797 \\
        NDA &+ Comma Stripping &0.809 \\
        \hline
        \end{tabular}
    \caption{Ablations to explore the impact of our dataset preprocessing. Comma stripping is stripping all commas that proceed a number, Lowercased is making the input and targets lowercase and Shuffled Epochs is reordering target values between every epoch (details in Section \ref{sec:entity-order}). Shuffled Epochs is not included for Kleister-Charity as there is a one to one mapping between key and value.}
    \label{tab:kleister-ablations}
\end{table}

\section{Limitations and Future Work}
\subsection{Tokenizer Limitations}
\label{section:standardization}
Although our work shows that sequence to sequence information extraction models are a viable alternative to token-level sequence-labeling tasks when database records are available for use as supervision, we hypothesize that T5's choice of tokenizer may make learning to jointly extract and standardize information more difficult than necessary.

\begin{figure}
    \centering
    \includegraphics[width=\columnwidth]{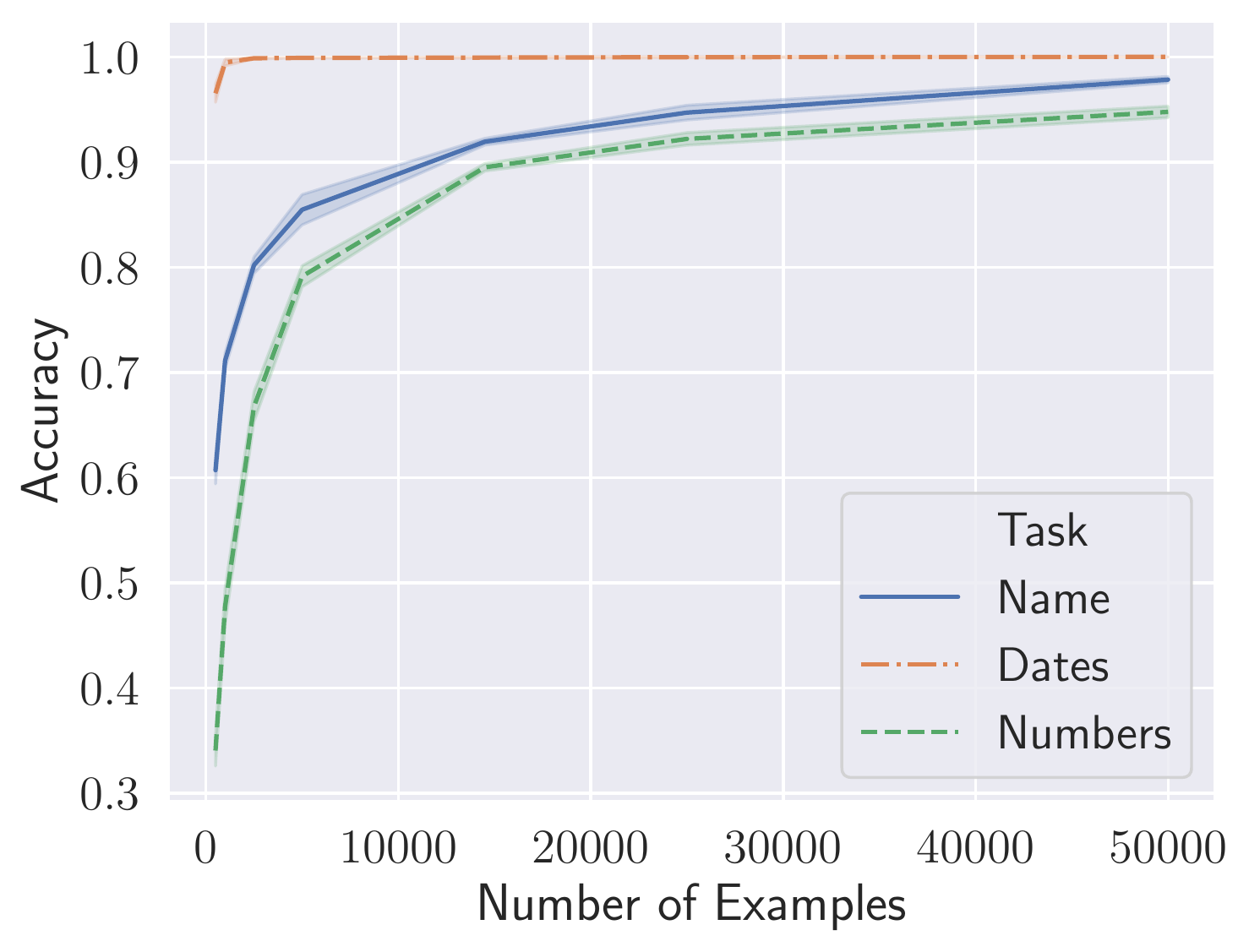}
    \caption{Name and number standardization is substantially more difficult than date standardization, which we attribute to tokenization differences caused by the standardization operation in these two tasks. See Section \ref{section:standardization} for experimental details.}
    \label{fig:standardization-results}
\end{figure}

In order to validate this hypothesis we opt to study the task of learning standardization rules in isolation to understand what kinds of standardization rules are most tractable to learn.  Below, we experiment with three synthetic datasets.  For all experiments we use the pre-trained T5 base checkpoint and measure how performance varies as dataset size is increased. Experiment results are presented in Figure \ref{fig:standardization-results}.

\begin{figure}
    \centering
    \includegraphics[width=\columnwidth]{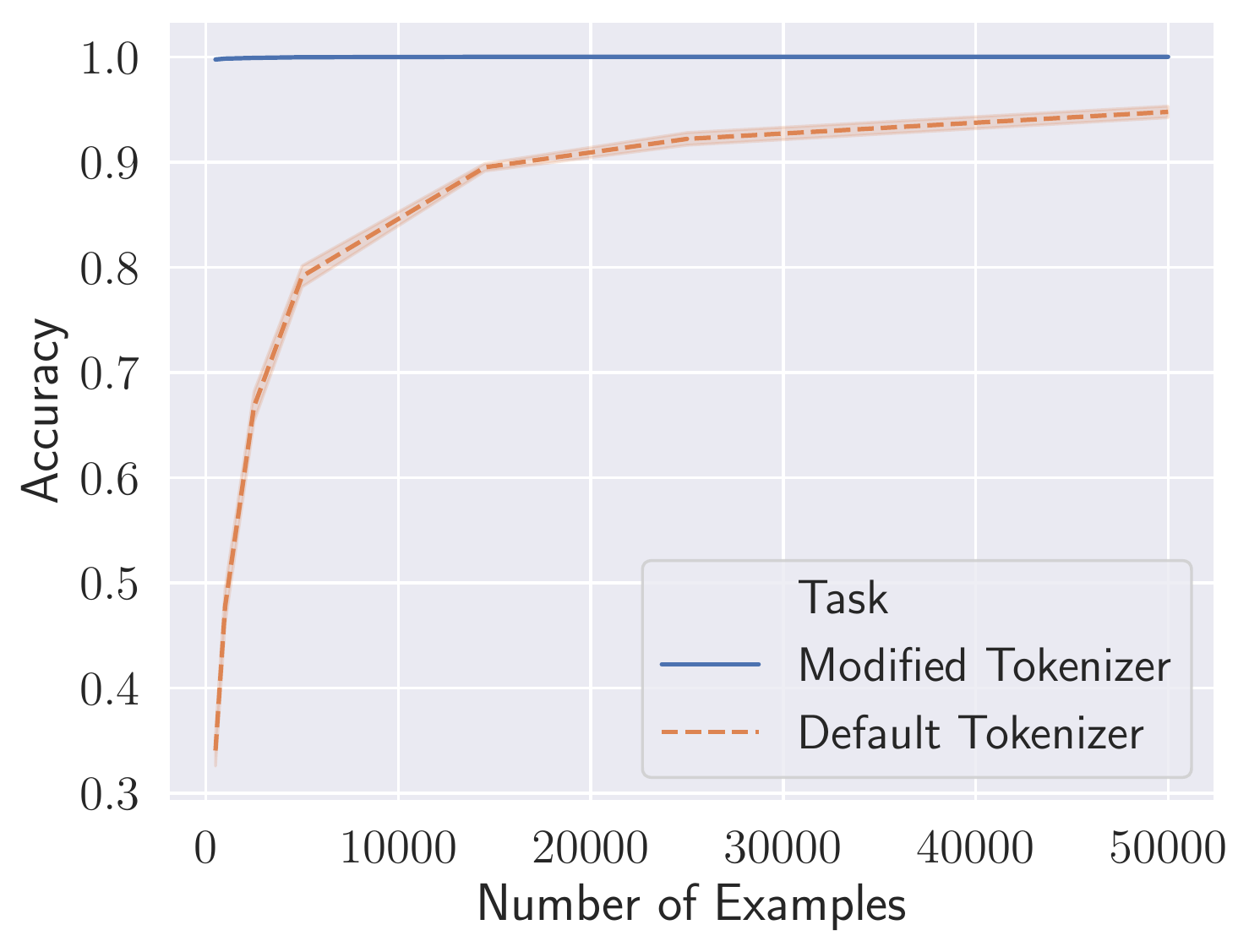}
    \caption{With T5's default tokenizer, we find learned standardization is imperfect even at 50k training examples. However, constraining T5's vocabulary to single character tokens makes the number standardization task trivial by ensuring tokens can be copied directly from the input.  Unfortunately, this change increases overall sequence length significantly, and we refrain from applying this modification in our main experimental results.}
    \label{fig:corrected-tokenizer}
\end{figure}

\subsubsection{Date Standardization}
The first synthetic dataset explores the difficulty in translating from a variety of date formats to a standard form. Random dates between 1950 and 2021 were generated using Python's datetime module and formatted using a family of 10 unique format strings. The model is tasked with translating these varied date formats to a canonical ``\%Y/\%m/\%d'' form.  As observed in our main results, we found this translation to be easily learnable -- the model completes the task with nearly perfect accuracy after only observing 2500 examples.

\subsubsection{Name Lowercasing}
The second synthetic dataset tests the model’s ability to learn to translate names with varying capitalization to a lowercase form. Of the input data, 5\% is lowercase, 20\% is uppercase, and the remaining 75\% has the first letter of every word capitalized.  Names are drawn at random from a gazetteer of roughly 18,000 names. In contrast to the date standardization task, at 2500 samples our exact match accuracy is only around 80\%.  Even at the full 50,000 samples we observe an error rate of 2.5\%, an indicator that this mapping is not trivial to learn. We hypothesize that this difficulty in learning to solve this task is due to the difference in the input and output sequence tokenization induced by the lowercasing operation (described previously in Section \ref{section:record-standardization}).

\subsubsection{Number Standardization}
Finally, we test the model’s ability to convert from integer values sampled uniformly between one thousand and one billion to a more human readable format that includes commas and two decimal points of precision.  A value of ``123456789'' would be converted to ``123,456,789.00'' -- the encoder-decoder model must learn to insert commas where appropriate and emit ``.00'' at the end of the sequence.  Similar to the lowercasing task, we observe a failure rate of ~5\% even when training on the full dataset of 50,000 samples.

\subsubsection{Tokenizer Modifications}
To validate our hypothesis that the poor performance on the number and name standardization datasets are due to the tokenization differences in the input and output sequences preventing a copying heuristic from being learned, we experiment with restricting the vocabulary of T5 to aid in copying number values.  In particular, we modify the tokenizer to produce only single character tokens for numeric values.  In Figure \ref{fig:corrected-tokenizer}, we show that this simple modification makes the standardization task trivial for the T5 model to solve, suggesting future work on tokenization may be key to making learned normalization practical.

\subsection{Conclusion}
\label{Limitations and Future Work}
We devise a new formulation for information extraction which directly exploits existing database records as training data, a much less labor-intensive approach than existing sequence-extraction methods. Paired with simple data preprocessing techniques and a strategy that enables processing long documents, our method can achieve comparable results to hand-crafted pipelines on several real-world datasets. 

Future work could incorporate constrained generation to avoid unparseable outputs, permutation-invariant losses to better address the issue of entity order, and information extraction-specific pretraining to help with learning implicit standardization. Moreover, as the fusion-in-decoder formulation does not allow for attention across chunks in the encoder, we hypothesize that methods which allow for long-range attention in the encoder during pre-training may lead to better task performance. Additionally, since document layout understanding is important for accurate information extraction (e.g. extraction of cell values in tables and forms), incorporating strategies to include the spatial location of tokens in their input representations, such as done in \cite{Xu2020LayoutLMPO}, could improve the general performance in our approach for information extraction. Finally, although not explored in this work, our flexible output format permits representing outputs with more complex structure and may prove a promising direction for approaching other information extraction tasks like relation extraction. Regardless, in its current form, Doc2Dict represents an easy, reproducible baseline for future information extraction tasks and paves the way towards practical information extraction processes.

\section{Acknowledgements}
We thank Luke Metz, Rahil Dedhia, Slater Victoroff and Matthew Bayer for their invaluable feedback on early drafts of this paper.

\bibliography{main}
\bibliographystyle{acl_natbib}

\newpage

\begin{appendix}

\section{Task Descriptions and Additional Experimental Details} 

\subsection{Kleister-NDA and Kleister-Charity}
% TODO: find a better title
For the Kleister datasets, we use the train, dev and test splits provided by the original paper and report uncased F1 scores \cite{gralinski2020}. Alongside the documents, the Kleister datasets provide outputs from several OCR providers. For each dataset we use the OCR provider that yielded the best metrics in the original paper: pdf2djvu for Kleister-NDA and Textract for Kleister-Charity. For each input sequence the Kleister dataset provides a set of "slots" to be filled; we concatenate this list of expected values to the beginning of the input sequence but do not enforce that responses for each slot are generated.

\paragraph{Kleister-NDA Statistics}
In order to validate the claim that heuristic alignment of normalized entities with the source text of the document leads to sub-par targets, we manually produce token-level annotations for a set of 100 Kleister-NDA documents.  We then attempt to automatically annotate the documents using a simple regular expression to search for near-exact matches in the source document. The matching procedure permits the inclusion of additional spaces, periods, or commas as well as variation in capitalization. Results of this evaluation are included in Table \ref{tab:kleister-nda-stats}.

For the ``Jurisdiction'' and ``Party'' labels we observe high recall, with all ``Jurisdiction'' labels and all but 7 ``Party'' labels having near-exact matches in the source document. Term and Effective Date labels are more varied in their presentation in the source document, with more than a fifth of relevant labels being missed by automatic annotation.  Perhaps more concerning, we also observe a high false positive rate for all classes, with precision scores varying from 0.25 to 0.72.  In other words, automatic annotations frequently match irrelevant mentions in the source document.

% Although we have not verified this hypothesis empirically, we hypothesize that these spurious matches could degrade model quality if left unfiltered.

\begin{table}[]
\centering
\caption{Statistics relating to fuzzy matching on the Kleister-NDA dataset. Automatically-labeled spans that overlap with ground truth annotations are considered true positives, those that do not are considered false positives.}\label{tab:kleister-nda-stats}
\begin{tabular}{lrrr}
\hline
&Recall & Precision\\
\hline
Effective Date &0.60 & 0.58\\
Jurisdiction & 1.00 & 0.33 \\
Party &0.94 & 0.42 \\
Term &0.72 & 0.72\\
\hline
\end{tabular}
\end{table}
% Kleister-Charity clearly poses some issues for our approach. While it is not clear exactly what these are we have been able to identify a number of possibly contributing factors.
\paragraph{Kleister-Charity Results}
We hypothesize that the poor performance of our approach on the Kleister-Charity task is due to several factors. First, documents in this task are significantly longer on average than the documents from other datasets (See Table \ref{tab:kleister-deepform-stats}.) Approximately 6.5\% of the documents in this task exceed our 32k token limit and are truncated.  This may require our model to produce a response for which it does not have access to supporting evidence.  Second, several of the Kliester-Charity fields commonly appear in tables (namely, "Income" and "Spending") instead of longer-form natural language. While this should also be a challenge for the baseline, we note that the chunking scheme used in Text2DB may exacerbate this issue by making it likely that a table cell value is not included in the same chunk as its corresponding header. 

\paragraph{Kleister Error Analysis}
We selected a sample of 20 documents from the Kleister-Charity and Kleister-NDA test sets in order to perform a qualitative error analysis. Overall, errors were similar to those seen in any information extraction pipeline.  We most commonly observed missed, erroneous, or partial matches akin to those produced by typical sequence-labeling models. 

We also occasionally observed errors unique to the text-generation problem framing. One such example are errors we attribute to a failure to copy a value from input to output. This seemed to be most prevalent in the Kleister-NDA post code field, where many of the extracted values were similar but not exactly the same as values that appeared in the document. In one case, we observed a field key predicted as a value in a case where the correct extracted value started with the same few tokens as a key: "Basic Income Earth Network (Bien)" became "basic income annually in british pounds". Another error mode unique to the text-generation framing is that of hallucinating plausible but invalid continuations of observed entities.  In one instance, our model extended the name of the town "Sutton" to "Sutton-upon-Thames".

It is worth noting that we observed no errors that could be described as "standardization errors", in cases such as dates and term where heavy transformations are applied by the model, the outputs were always correctly formed.

\subsection{DeepForm}
The DeepForm dataset is composed of PDF disclosures of advertising expenditures from US political campaigns. The ground truth for the dataset includes values which have been manually recorded in an effort to make the political process more transparent. Information about the location on the page where entities were found is not available, and existing benchmarks on the task employ a complex fuzzy matching process that attempts to align extracted values with candidate tokens from the source documents. We selected the labels "Gross Amount", "Committee", "Agency" and "Call Sign" to include in our experiments. DeepForm does not provide any standardised splits for the data, and because the source documents are continuously pulled from a 3rd party website, the documents that are included in the dataset can vary depending on the time in which the dataset is downloaded. For this reason we run our own baselines for DeepForm. We use the pdf-to-text pipeline and fuzzy matching procedure from the DeepForm codebase and finetune a RoBERTa model \cite{Liu2019} with a conditional random field output layer as our baseline. At prediction we select the most confident candidate prediction. Additionally, we write simple output standardization rules for gross amount to remove any leading dollar signs and to convert the numbers back into the format used for input. 

\subsection{OntoNotes 5.0}

For OntoNotes, we attempt to reproduce the train, dev and test splits as reported in \citet{chiu-nichols-2016-named} and \citet{ghaddar-langlais-2018-robust}. To this end, we utilized the train and dev splits as produced by the script provided in the CoNLL-2012 task \cite{pradhan-etal-2012-conll}, and the test split as produced by the script provided alongside \citet{pradhan-etal-2013-towards}. Finally, following \citet{ghaddar-langlais-2018-robust} we exclude the New Testament portion, as it does not contain gold standard NER labels.

\begin{table*}[]
    \centering
    \begin{tabular}{lrrrrrrr}\hline
        &Beam Size & Epochs & Chunks & Lowercase & Shuffled Epochs & Strip Commas \\\hline
        NDA &2 &128 &32 &Yes &Yes &Yes \\
        Charity &4 &32 &64 &Yes &- &Yes \\
        DeepForm &4 &8 &16 &Yes &- &Yes \\
        DeepForm Baseline &- &8 &- &- &- &- \\
        \hline
    \end{tabular}
    \caption{Best hyper parameters for each run. Learning rate was always $6.25e-5$ and batch size always $1$.}
    \label{tab:kleister-deepform-hparams}
\end{table*}

\begin{table*}[]
    \centering
    \begin{tabular}{llrr}\hline
        Dataset &Label & Doc2Dict F1 & Baseline F1 \\
        \hline
        DeepForm &Gross Amount &0.853 &\textbf{0.98} \\
        DeepForm &Comittee &\textbf{0.844} &0.615 \\
        DeepForm &Agency &\textbf{0.907} &0.721 \\
        DeepForm &Callsign &\textbf{0.996} &0.727 \\
        Deepform & &\textbf{0.9} &0.761 \\
        \hline
        NDA &Party &\textbf{0.761} &0.701 \\
        NDA &Juristiction &\textbf{0.961} &0.938 \\
        NDA &Effective Date &\textbf{0.836} &0.82 \\
        NDA &Term &0.56 &\textbf{0.608} \\
        NDA & &\textbf{0.809} &0.777 \\
        \hline
        Charity &Street &0.571 &\textbf{0.682} \\
        Charity &Charity Name &0.665 &\textbf{0.724} \\
        Charity &Income &0.168 &\textbf{0.709} \\
        Charity &Postcode &0.580 &\textbf{0.836} \\
        Charity &Charity Number &0.827 &\textbf{0.967} \\
        Charity &Town &0.659 &\textbf{0.833} \\
        Charity &Report Date &0.932 &\textbf{0.957} \\
        Charity &Spending &0.124 &\textbf{0.685} \\
        Charity & &0.561 &\textbf{0.801} \\
        % 64 epochs, batch size 8
    \end{tabular}
    \caption{Full Per-field Results for Kleister-NDA, Kleister-Charity and DeepForm.}
\end{table*}

\begin{table*}[]
    \centering
    \begin{tabular}{lrrrrrr}
        \hline
        \textbf{Dataset} & \textbf{Split} & \textbf{Num Docs} & \textbf{Min Length} & \textbf{Max Length} & \textbf{Mean length} \\
        \hline
        DeepForm &Train &5981 &272 &114994 &3715 \\
        DeepForm &Test &2567 &268 &93450 &3676 \\
        \hline
        NDA &Train &254 &817 &26109 &3802 \\
        NDA &Dev &83 &494 &10418 &4070 \\
        NDA &Test &203 &751 &14074 &3802 \\
        \hline
        Charity &Train &1729 &509 &96669 &12246 \\
        Charity &Dev &440 &379 &122258 &13950 \\
        Charity &Test &609 &520 &365630 &14272 \\
    \hline
    \end{tabular}
    \caption{Dataset statistics measures in tokens of T5's tokenizer.}
    \label{tab:kleister-deepform-stats}
\end{table*}

\end{appendix}
\end{document}